\documentclass[runningheads]{llncs}

\usepackage{graphicx}
\usepackage{comment}
\usepackage{amsmath,amssymb}

\usepackage{xcolor}
\usepackage{url}
\usepackage{hyperref}

\usepackage{caption}
\usepackage{subcaption}

\newcommand{\orcidhack}[1]{\href{https://orcid.org/#1}{\includegraphics[width=7pt]{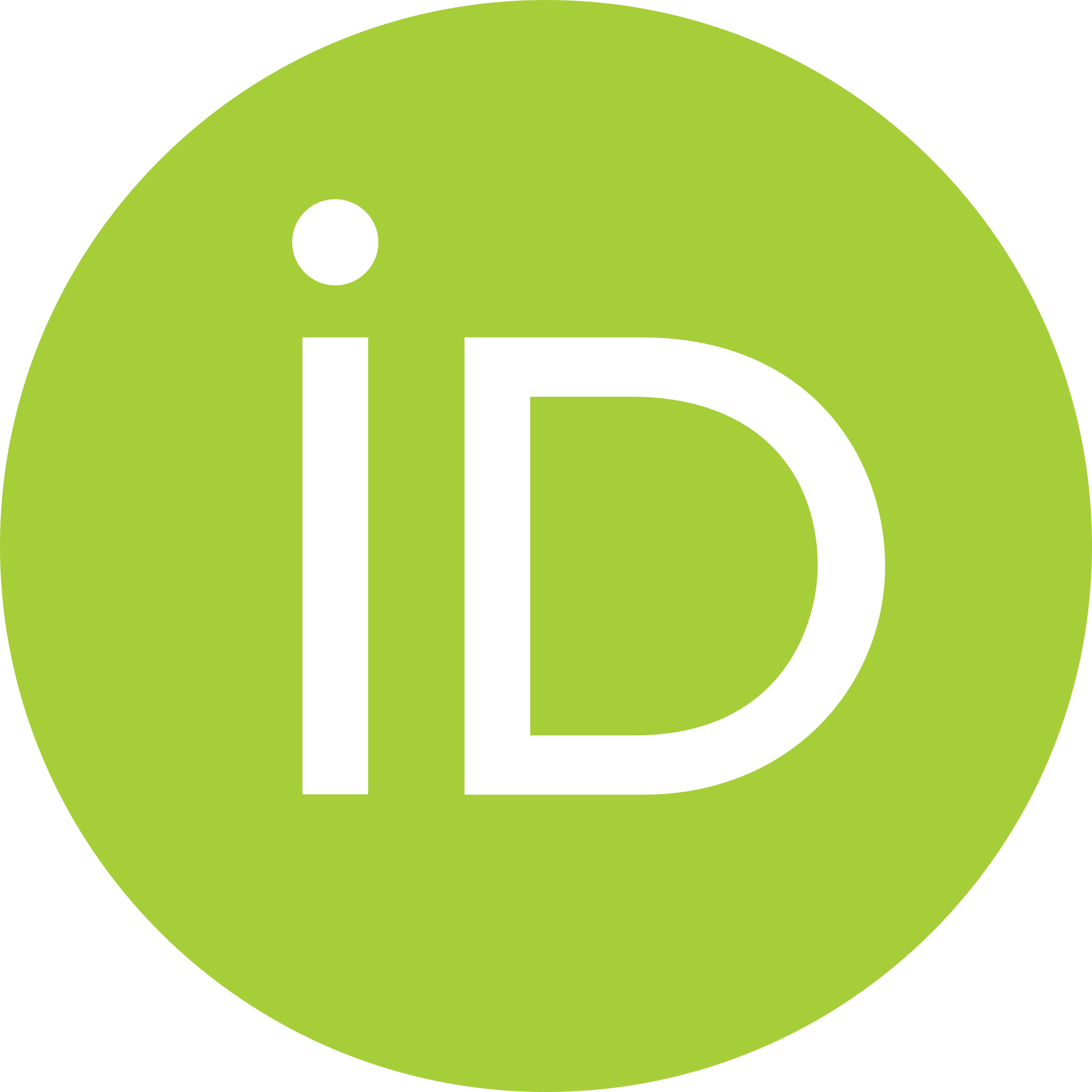}}}

\newif\ifreview
\reviewfalse

\ifreview
	\usepackage{lineno}

	\linenumbers
\fi

\begin{document}


\def\SubNumber{061}

\def\GCPRTrack{Main Track}

\title{Hyperspectral Demosaicing of Snapshot Camera Images Using Deep Learning}

\ifreview
\else
	\titlerunning{Hyperspectral Demosaicing of Snapshot Camera Images}

	\author{Eric L.~Wisotzky\inst{1,2} \orcidhack{0000-0001-5731-7058} \and
	Charul Daudkhane\inst{1} \and
	Anna Hilsmann\inst{1} \orcidhack{0000-0002-2086-0951} \and
	Peter Eisert\inst{1,2} \orcidhack{0000-0001-8378-4805}}
	
	\authorrunning{E. Wisotzky et al.}
	
	\institute{Computer Vision \& Graphics, Fraunhofer Heinrich Hertz Institute, Berlin, Germany \and
	Visual Computing, Institut f\"ur Informatik, Humboldt-University, Berlin, Germany\\
	\email{eric.wisotzky@hhi.fraunhofer.de}}
\fi

\maketitle              

\begin{abstract}
Spectral imaging technologies have rapidly evolved during the past decades. The recent development of single-camera-one-shot techniques for hyperspectral imaging allows multiple spectral bands to be captured simultaneously ($3\times 3$, $4\times 4$ or $5\times 5$ mosaic), opening up a wide range of applications. Examples include intraoperative imaging, agricultural field inspection and food quality assessment. To capture images across a wide spectrum range, i.e.~to achieve high spectral resolution, the sensor design sacrifices spatial resolution. With increasing mosaic size, this effect becomes increasingly detrimental. Furthermore, demosaicing is challenging. Without incorporating edge, shape, and object information during interpolation, chromatic artifacts are likely to appear in the obtained images. Recent approaches use neural networks for demosaicing, enabling direct information extraction from image data. However, obtaining training data for these approaches poses a challenge as well. This work proposes a parallel neural network based demosaicing procedure trained on a new ground truth dataset captured in a controlled environment by a hyperspectral snapshot camera with a $4\times 4$ mosaic pattern. The dataset is a combination of real captured scenes with images from publicly available data adapted to the $4\times 4$ mosaic pattern. To obtain real world ground-truth data, we performed multiple camera captures with 1-pixel shifts in order to compose the entire data cube. Experiments show that the proposed network outperforms state-of-art networks.

\end{abstract}
\section{Introduction}
Spectral imaging technologies have rapidly evolved during the past decades. The light captured by the camera sensor contains information across a wide range of the electromagnetic spectrum, which can potentially be used to identify the texture and the material / molecular composition of any given object of interest. Hyperspectral sensors with up to 150 or even 256 spectral channels within and outside the visible spectral range have been developed in the last few years for various applications within health care \cite{calin2014hyperspectral,Lu2014}, industrial imaging \cite{shafri2012hyperspectral} or agriculture \cite{jung2006hyperspectral,moghadam2017plant}. However, current acquisition methods for such devices (as filter-wheels \cite{Wisotzky2018}, line-scanning \cite{WisotzkyComparision2021}) have decisive disadvantages, trading the acquisition of hyperspectral images (HSI) with high costs or long acquisition times. Alternative approaches are Multi Spectral Filter Arrays (MSFA) based on spectral masking on pixel-level using a single sensor plane similar to the Bayer pattern for single chip RGB cameras \cite{bayer1976color}. This single-camera-one-shot (mosaic snapshot) technique allows multiple spectral bands to be captured simultaneously in a simple and compact system. The captured image data is defined by its moxel (mosaic element) corresponding to the occurring filter pattern and stored in a hypercube representation with three dimensions, two spatial (width and height) and one spectral (wavelength $\lambda$). The simplest example is similar to the Bayer pattern, with one green filter element being replaced by another filter resulting in a $2\times2$ pattern \cite{hershey2008multispectral}. These systems can be extended to $3\times3$, $4\times4$, $5\times5$ or even non-quadratic $2\times3$ mosaic patterns for recording wavelengths in near-ultraviolet, visible and near-infrared spectral range. However, a mosaic pattern is always a compromise between spatial and spectral resolution, as with increasing mosaic size (for higher spectral resolution) the spatial resolution decreases. This spectral-spatial trade-off can be resolved through interpolation or prediction of the missing spectral values, such that the final HSI exhibits higher resolution (spatial and spectral). Techniques for spectral reconstruction include bilinear and nonlinear filtering methods and are referred to as demosaicing.

Typically, demosaicing is achieved by interpolation based on the information from neighboring pixels. Traditional algorithms such as bilinear, bicubic interpolation are popular choices in the field of image processing, where missing pixels are calculated from their neighborhood. Further, interpolation methods based on image fusion have been proposed, where spatial-resolution is enhanced through fusion with Pseudo Panchromatic Images (PPI), statistics based fusion techniques, e.g.~Maximum a posteriori (MAP) estimation, stochastic mixing model \cite{eismann2004application,eismann2005hyperspectral,hardie2004map}, and dictionary-based fusion techniques, e.g.~spectral and spatial dictionary based methods \cite{bendoumi2014hyperspectral,yokoya2011coupled,zhang2014spatial}. 
Fusion based methods usually require the availability of a guiding image with higher spatial resolution, which is difficult to obtain in many scenarios.
Demosaicing by interpolation based techniques, both traditional as well as fusion-based, is easy to achieve, however, these methods suffer from color artifacts and lead to lower spatial resolution. 
Especially at edges, they do not take into account the spectral correlations between adjacent bands as well as due to crosstalk. This results in spectral distortions in the demosaiced image, especially for increasing mosaic filter size.

Alternatively, deep neural networks can be trained to account for scene information as well as correlations between individual spectral bands.
Demosaicing using convolutional neural networks (CNN) for images with $2\times2$ Bayer pattern was first proposed in \cite{wang2014multilayer} and \cite{gharbi2016deep}. In recent years, CNN based color image super resolution (SR) has gained popularity. Examples of such networks include SRCNN \cite{dong2014learning}, DCSCN \cite{yamanaka2017fast} and EDSR \cite{lim2017enhanced}. Due to their success, these networks have been extended to HSI super resolution \cite{li2018single}. The underlining aspect of all CNN based HSI demosaicing networks is the utilization of spatial and spectral context from the data during training.
However, the need of high quality ground truth data leads to challenges. In such a dataset, each pixel should contain the entire spectral information, which is difficult to acquire in a natural environment.  

The contribution of this work is as follows. We present a new ground-truth dataset acquired and generated using a shifting unit to achieve a 1-pixel movement on the camera side in order to obtain a full resolution image for all color channels. Further, we propose a new demosaicing network and compare it to three relevant network architectures, performing demosaicing on a dataset combining captured and publicly available data.

The remainder of this paper is as follows. The next chapter gives an overview on related publications relevant for this work. Chapter 3 describes the proposed network architecture, before chapter 4 explains the acquisition of the ground truth data. Chapter 5 introduces training and evaluation parameters. Chapter \ref{sec:res} describes experiments and results, followed by a thorough discussion and conclusion.

\section{Related Work}
In the last years, several HSI demosaicing algorithms have been presented. Some methods require presence of a dominant-band (as in the Bayer pattern) \cite{monno2012multispectral}, but of more interest are methods without such explicit assumptions designed for mosaic pattern having no redundant band \cite{ogawa2016demosaicking}. 

Dijkstra et al.~\cite{dijkstra2019hyperspectral} proposed a similarity maximization network for HSI demosaicing, inspired by single image SR. This network learns to reconstruct a downsampled HSI by upscaling via deconvolutional layers. The network results are presented for a $4\times4$ mosaic pattern and the demosaiced HSI showed high spatial and spectral resolution. Habtegebrial et al.~\cite{habtegebrial2019deep} use residual blocks to learn the mapping between low and high resolution HSI, inspired by the HCNN+ architecture \cite{shi2018hscnn+}.
These two networks use 2D convolutions in order to learn the spectral-spatial correlations. An important characteristic of HSI is the correlation between adjacent spectral bands which are not taken into account when using 2D convolutional based networks. These correlations can be incorporated by using 3D convolutional networks. Mei et al.~\cite{mei2017hyperspectral} proposed one of the first 3D CNNs for hyperspectral image SR addressing both the spatial context between neighboring pixels as well as the spectral correlation between adjacent bands of the image. A more effective way to learn the spatial and spectral correlations is through a mixed 2D/3D convolutional network, as proposed in \cite{li2020mixed}. 

One major challenge for the task of snapshot mosaic HSI demosaicing using neural networks is the lack of real world ground truth data. Publicly available real world datasets, such as CAVE \cite{yasuma2010generalized} or HyTexiLa \cite{khan2018hytexila}, were recorded either using a push-broom technique or by spatio-spectral-line scans. 
Hence, the data has different characteristics than snapshot mosaic data (e.g., missing cross talk) and can therefore not be used to adequately train a robust network for demosaicing. One alternative is a downsampling strategy from captured snapshot mosaic data as presented in \cite{dijkstra2019hyperspectral}. However, simple downsampling leads to differences in distances of adjacent pixels, which affects the network results.

\section{Network Architecture}
\begin{figure}[htp]
  \centering
  \includegraphics[width=0.2\columnwidth]{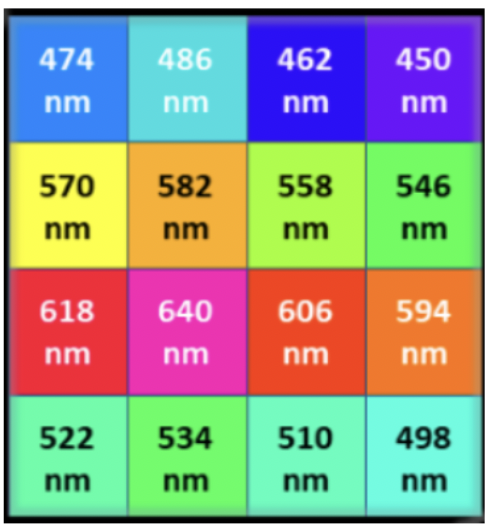}
  \caption{ $4\times 4$ filter pattern with 16 wavelengths}
	\label{fig:filter_pattern}
\end{figure}

We propose a new neural network architecture to generate a full-spectrum hypercube (dimension $[L \times W \times H]$) with $L$ wavelength bands from a captured mosaic image (dimension $[1\times W \times H]$), represented as a low resolution or sparse hypercube. Here, $W$ and $H$ correspond to the spatial width and height of the image and $L$ represents the spectral resolution. In this work, a snapshot mosaic camera with a $4\times4$ pattern, i.e.~$L = 16$ bands, is used as shown in Fig.\ref{fig:filter_pattern}.

\begin{figure}[t]
  \centering
	\subfloat[Full network\label{fig:network_general}]{\includegraphics[width=0.4\columnwidth]{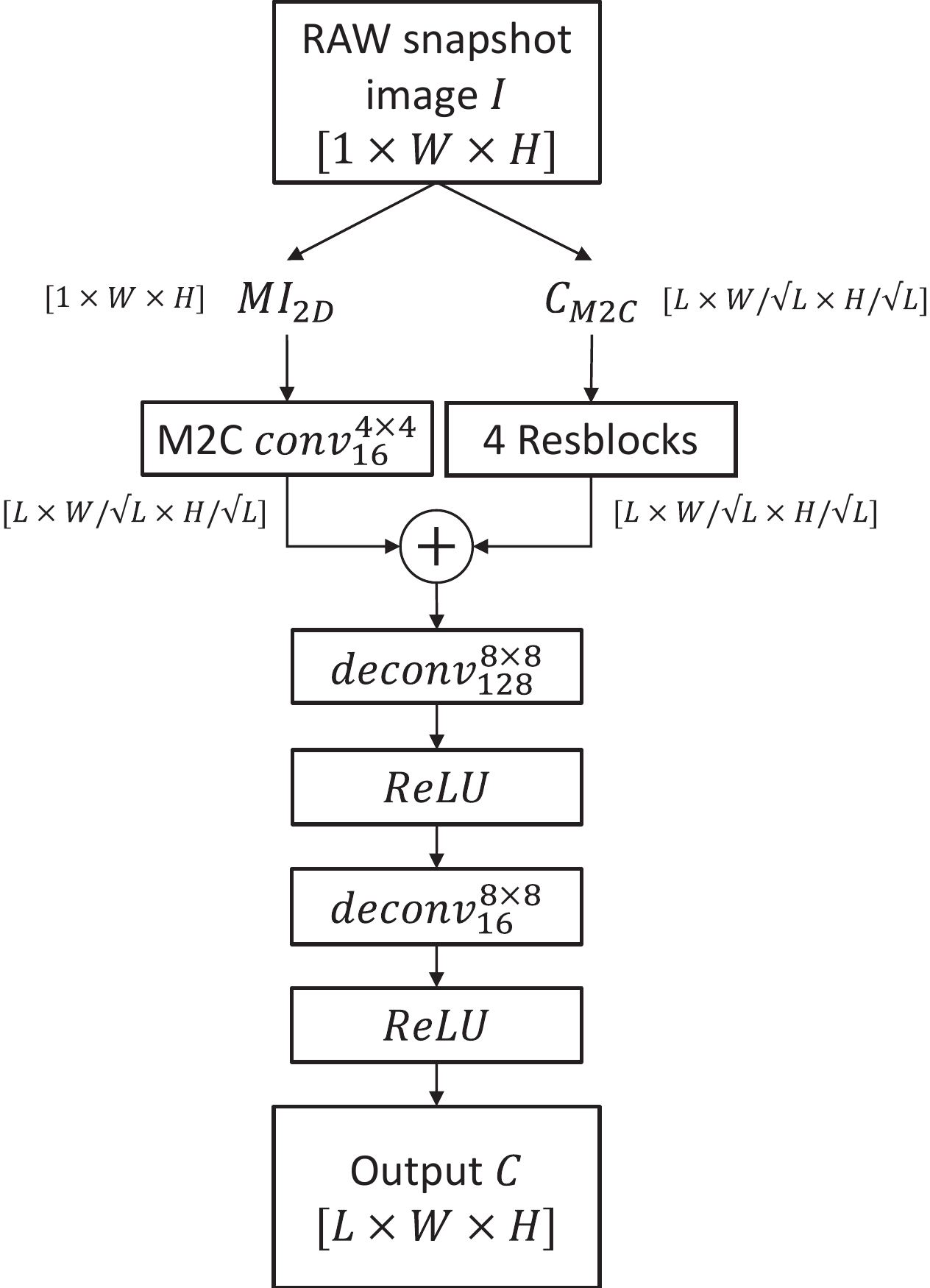}}\
	\subfloat[Resblock\label{fig:network_resnet}]{\includegraphics[width=0.25\columnwidth]{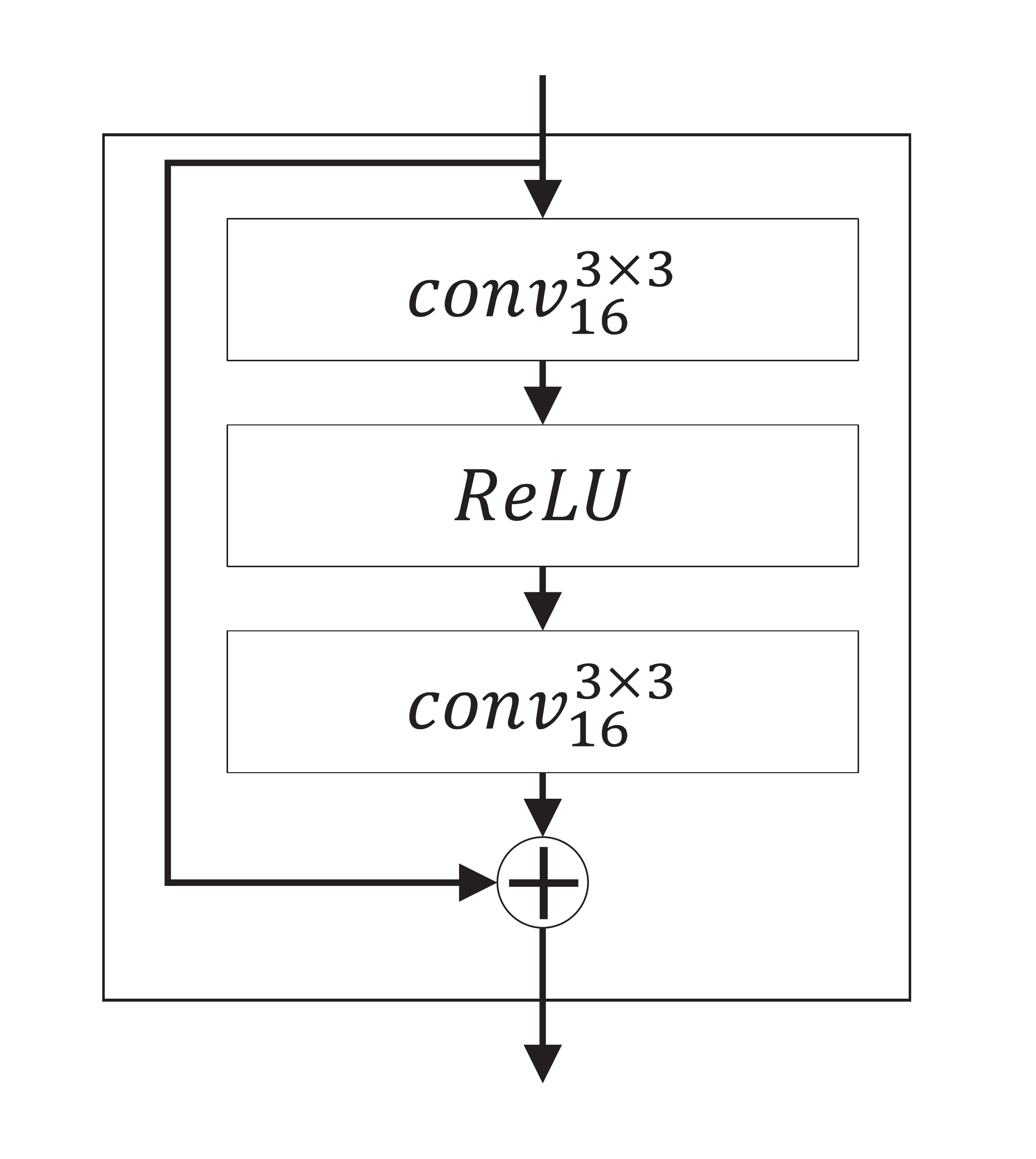}}
  \caption{(a) Architecture of the hyperspectral demosaicing network. (b) Architecture of the resnet block. The full network includes four consecutive resnet blocks. The resnet block is build up of two convolutional layers holding $16$ convolutional filters with size $3\times 3$ each. These two layers are separated by a ReLU activation function.}
\end{figure}

The building blocks of our demosaicing network (Fig.~\ref{fig:network_general}) are a feature extraction layer along with a feature addition (\textit{FeatureAdd}) block and two deconvolution (\textit{deconv}) layers. The feature extraction layer of the network is split into two parallel parts: (1) a mosaic to cube converter (M2C) implemented by a convolutional layer (\textit{conv}) and (2) feature extraction using four residual blocks (\textit{Resblocks}). These two paths learn the spatial spectral correlations from different representations of the input mosaic image. The input to the M2C is a 2D mosaic image $MI_{2D}$ of dimension $[1, W, H]$, while the input to the \textit{Resblocks} is a 3D cube representation $C_{M2C}$ of dimension $[L, W/\sqrt{L}, H/\sqrt{L}]$ created by resampling the one channel input image as follows: 
\begin{equation}
C_{M2C}(x,y,z) = MI_{2D}(u, v),
\label{eq:M2Chandcraft}
\end{equation}
with
\begin{align}
u &= x \cdot \sqrt{L} + z\ \text{mod}\ \sqrt{L},\\
v &= y \cdot \sqrt{L} + z\ \text{div}\ \sqrt{L},
\end{align}
where $u, v$ correspond to the 2D mosaic pixel coordinate, $x, y, z$ correspond to the 3D multispectral voxel coordinate and $\sqrt{L}$ is the dimension of the mosaic pattern; in our case $\sqrt{L} = 4$. The operation $\text{div}$ describes integer division. The M2C composed of a convolutional layer is defined as
\begin{equation}
C_{nn}(x,y,z) = MI_{2D} \otimes_4 \mathcal{G}^{4\times 4}_{16},
\label{eq:M2Cconv}
\end{equation}
where $\otimes$ and $\mathcal{G}$ represent the convolutional operator with stride of $4$, equal to mosaic size, and the set of $16$ filters with a size of $4\times 4$ respectively.
The features of $C_{M2C}$ are extracted by four consecutive residual blocks, where a single block consists of two convolutional layers separated by a rectified linear unit (ReLU) activation function and interconnected through skip connection avoiding the problem of vanishing gradients \cite{hochreiter1998vanishing}. In addition, the ReLU function clips values below zero as negative spectral responses cannot exist. Each convolutional layer uses a filter of dimension $3\times 3$ with stride and padding of $1$, see Fig.~\ref{fig:network_resnet}. Finally, a ReLU activation function is applied to the obtained feature map and the resulting feature map is passed onto the next residual block as input.

The extracted features are of the same size and concatenated in the FeatureAdd block. The combined feature map is passed to two upsampling layers, which upsample by a factor of $4\times 4 = 16$ to produce a fully defined hyperspectral cube of dimension $[L, W, H]$ in a non-linear fashion. The first deconv layer determines the capacity of the network with $128$ filters, while the second layer has the same amount of filters as the number of required spectral bands
\begin{equation}
O(C) = \Phi(\Phi(C_{add} \oslash_2 \mathcal{F}^{8\times 8}_{128}) \oslash_2 \mathcal{F}^{8\times 8}_{16}),
\label{eq:Upsampling}
\end{equation}
where $\oslash$ and $\mathcal{F}$ represent the deconvolutional operators with stride of $2$ and the filter sets with a size of $8\times 8$ respectively. The ReLU activation function $\Phi$ after each \textit{deconv} layer accounts for non-linearity in the interpolation process.

\section{Data Acquisition}
In order to train the network with real ground truth images, we created a customized dataset from publicly available as well as new self-recorded ground truth data. 

A HSI captured with a $4\times 4$ mosaic pattern contains information for $1$ out of $16$ wavelength bands in each pixel only. For training the demosaicing algorithm, ground-truth data is necessary such that for each pixel, full spectral information exists across all $16$ wavelength bands. Hence, we captured an unprecedented new HSI dataset, providing accurate ground truth upsampling information. We capture in an controlled environment with a Ximea snapshot 4x4-VIS camera using an IMEC CMOS sensor. The sensor resolution is $2048\times 1088$ px, with an active area of $2048\times 1024$ px and the $4\times 4$ mosaic filter pattern captures $16$ wavelength bands in the spectral range from $463$ nm to $638$ nm. 

For the generation of the ground-truth information, we captured $16$ images with precise pixel-wise shifting, meandering along the $4\times 4$ mosaic pattern for each scene, using a computer-controlled shifting unit. To achieve accurate 1-pixel shifts, the camera was calibrated and the scene-camera distance was measured. After acquiring the $4\times 4 = 16$ images, a full resolution image with $L = 16$ wavelengths was created by resampling the image stack.
At each position, we captured and averaged several images to decrease the influence of sensor noise. In order to account for illumination differences and work with reflectance intensity, all captured images were corrected using a white reference image according to \cite{Wisotzky2019SPIEMI}. Further, snapshot mosaic images contain spectral crosstalk, which influences the reflectance behavior \cite{Wisotzky2020JMI}. To learn this behavior during demosaicing, the ground-truth data was crosstalk corrected in addition to white reference correction, while the training and test input data were white reference corrected only.

\begin{figure}[!ht]
  \centering
  \subfloat[Experimental setup with a color chart as captured object]{\includegraphics[width=0.7\columnwidth]{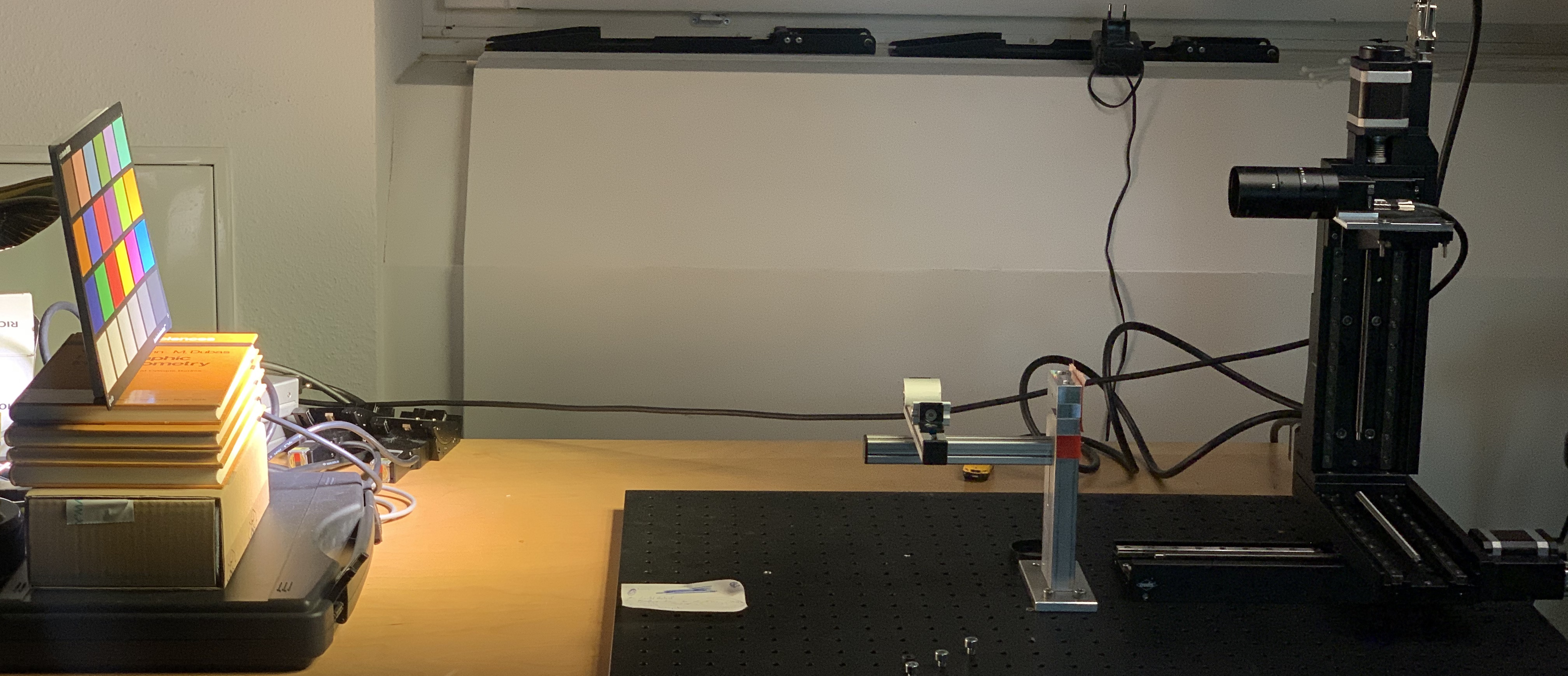}}
	\\
	\subfloat[Scene 2]{\includegraphics[width=0.394\columnwidth]{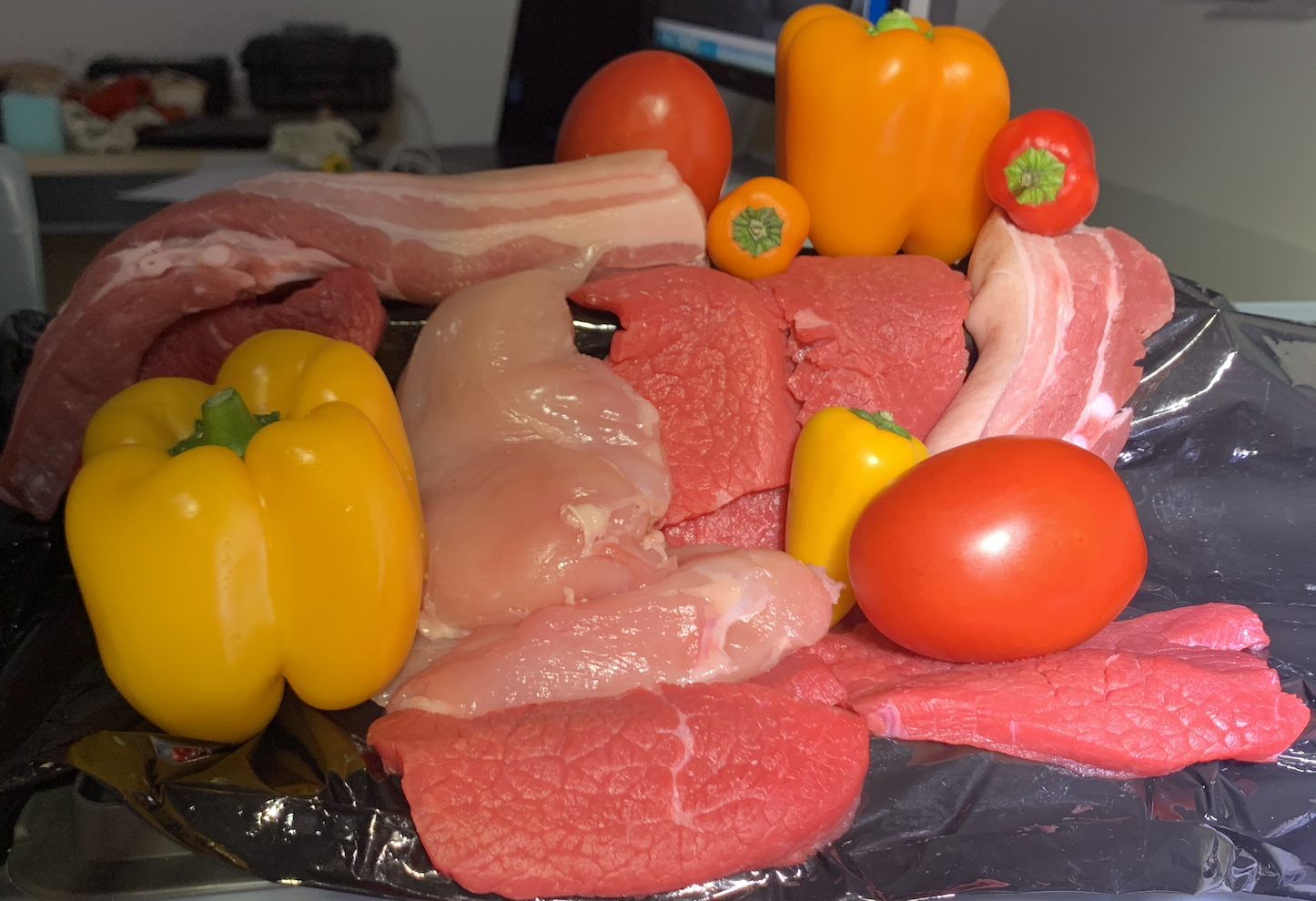}}\
	\subfloat[Scene 3]{\includegraphics[width=0.405\columnwidth]{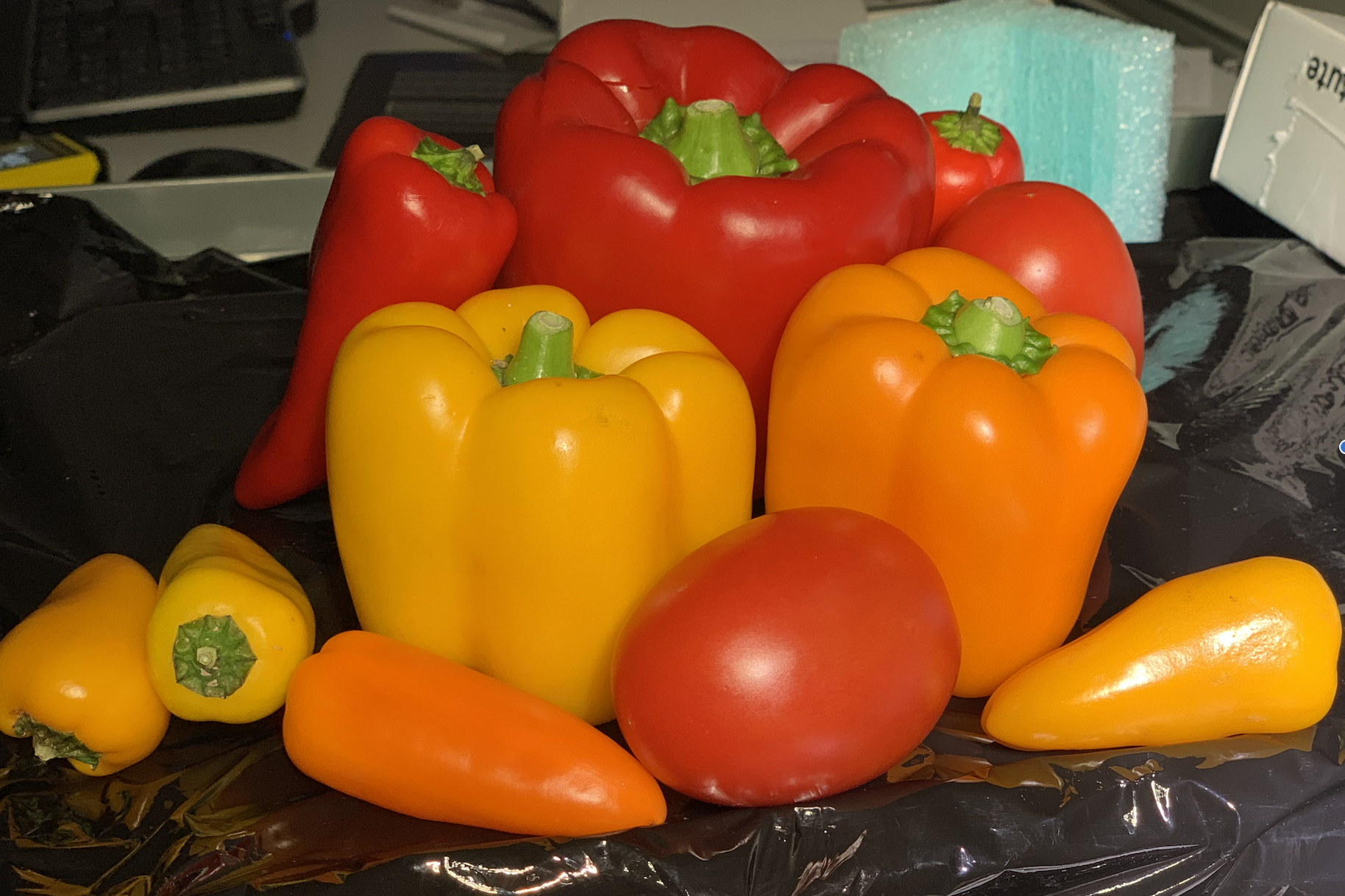}}
  \caption{(a) Experimental setup for recording hyperspectral images using a computer-controlled shifting unit, allowing exact 1-px shifts of the camera. Examples for two captured scenes containing (b) a combination of vegetables and meat and (c) different types of vegetables.}
	\label{fig:experimental_setup}
\end{figure}

We captured a total of twelve scenes, six with a color chart and two of each showing different vegetables, different meats, as well as a combination of vegetables and meats, see Fig.~\ref{fig:experimental_setup}. 
It is important to note, that a correct 1-pixel shift can only be achieved for the focus plane and points in front or behind that plane will always show slightly divergent pixel shifts of ($1\text{-px}\pm \epsilon$). For this reason, we added a flat color chart to half of the entire dataset and filtered the remaining images with a Gaussian filter with $\sigma = 1.5$ to smooth the present shifting error.

In addition, to increase the dataset, our captured data was combined with synthetic images generated from the publicly available CAVE dataset \cite{yasuma2010generalized}. The CAVE dataset has 32 reflectance images of real-world objects. In total $18$ images of the CAVE data sections 'skin and hair', 'food and drinks', 'real and fake' as well as 'stuff' were added to our dataset by interpolating the needed spectral bands and building up a simulated mosaic pattern representation of each scene.
The entire dataset will be available upon publication.

\section{Training and Network Evaluation Metrics}
During training, we used image patches of size $[1\times 100\times 100] = [16\times 25\times 25]$ as input for the network. We split the dataset into a training set with $1000$ patches of eleven captured images and 17 images of the CAVE data as well as a test set including $75$ patches of each of the two datasets.
We used the ADAM optimizer with an adaptive learning rate strategy and an initial learning rate of $0.001$ \cite{kingma2014adam}. The learning rate was reduced after each epoch until a value of $0.0001$ was reached, using at least $30$k epochs. A batch size of $20$ was used for updating the weights during the network training. The M2C block was initialized with uniform random weight. The loss function for calculating the difference between the ground truth and the predicted full-spectrum hyperspectral cube is defined by the mean squared error (MSE)
\begin{equation}
MSE(o,p) = \frac{1}{N} \sum_{i=0}^{N}{|o_i-p_i|^2},
\end{equation}
where $o$ is the ground truth and $p$ is the predicted value.

\section{Experiments and Results} \label{sec:res}
For quantitative analysis, we use the structural similarity index (SSIM), measuring the similarity between spectral cubes \cite{wang2004image} from predicted and ground-truth data as well as the peak signal to noise ratio (PSNR) \cite{hore2010image}.
Both, SSIM and PSNR are calculated individually for each spectral channel and averaged over all channels for the test images.

We compare the results of our proposed network to two state-of-the-art demosaicing approaches DCCNN \cite{dijkstra2019hyperspectral} and DeepCND \cite{habtegebrial2019deep}. Further, we analyze the resulting images visually and, to show the usability of our work, we visually analyze intraoperative snapshot images acquired during a parotidectomy. 

\subsection{Quantitative Results}
\begin{table}[b]
\caption{SSIM and PSNR results of our proposed network and the DCCNN \cite{dijkstra2019hyperspectral} using $32$ and $128$ filters.}
\label{tab:res_filter}
\centering
\begin{tabular}{l|cc|cc|r}
\hline
\# of filters & \multicolumn{2}{c}{32} & \multicolumn{2}{c}{128} & \\
Networks 			& SSIM & PSNR & SSIM & PSNR & epochs \\
\hline
DCCNN \cite{dijkstra2019hyperspectral} & 0.755 & 40.05 & 0.825 & 41.54 & $30$k \\
Ours					& 0.776 & 40.50 & 0.836 & 42.11 & $30$k \\
Ours					& 0.784 & 40.95 & 0.841 & 42.63 & $40$k \\
\hline
\end{tabular}
\end{table}
All networks, our proposed network as well as the two reference networks, were trained on the created dataset and the networks learned to predict a full spectral cube of dimension $[16\times 100\times 100]$ from the input mosaic image of dimension $[1\times 100\times 100]$. As the quality of the results depends on the number of filters in the first deconvolutional layer, with best results reported between $32$ and $256$ filters \cite{dijkstra2019hyperspectral}, we trained two versions of our as well as the DCCNN \cite{dijkstra2019hyperspectral} network using $32$ and $128$ filters in the first deconvolutional layer. Table \ref{tab:res_filter} reports the SSIM and PSNR results of our network and the DCCNN using $32$ and $128$ filters, respectively. Using $32$ filters and $1000$ images as well as $30$k epochs for training, the DCCNN network shows results comparable to the initial results reported in Dijkstra et al.~\cite{dijkstra2019hyperspectral}. Our proposed network outperforms the reference network approximately by $2\%$ using the same parameter and training set, see Tab.~\ref{tab:res_filter}. Further, the results show that using $128$ filters yields better results as with $32$ filters as recommended by Dijkstra et al.~\cite{dijkstra2019hyperspectral}. Therefore, for further analysis, we have used $128$ filters in the first \textit{deconv} layer for the other experiments.

\begin{table}[t]
\caption{PSNR results of all analyzed demosaicing methods. }
\label{tab:res_final}
\centering
\begin{tabular}{l|c}
\hline
Networks                & PSNR [dB] \\
\hline
Ours                    & 43.06 \\
DeepCND \cite{habtegebrial2019deep} & 42.38 \\
DCCNN \cite{dijkstra2019hyperspectral} & 41.54 \\
2D-3D-Net \cite{Charul21} & 41.33 \\
Intensity Difference \cite{Mihoubi2015} & 40.43 \\
Bilinear Interpolation  & 40.23 \\
Bicubic Interpolation   & 39.20 \\
\hline
\end{tabular}
\end{table}

In addition, we noticed that with $30$k epochs, the training error was still decreasing for our network, while it converged for the reference networks. Therefore, the quality of our results increases further when using up to $40$k epochs during training. The final results, shown in Tab.~\ref{tab:res_final}, present the PSNR at lowest training loss showing that our model outperforms the two reference networks as well as traditional interpolation methods by approximately $1$ dB (compared to DeepCND \cite{habtegebrial2019deep}) to $4$ dB (compared to bicubic interpolation).

\subsection{Qualitative Results}
\begin{figure}[b]
  \centering
  \subfloat[Ground Truth Image]{\includegraphics[width=0.49\columnwidth]{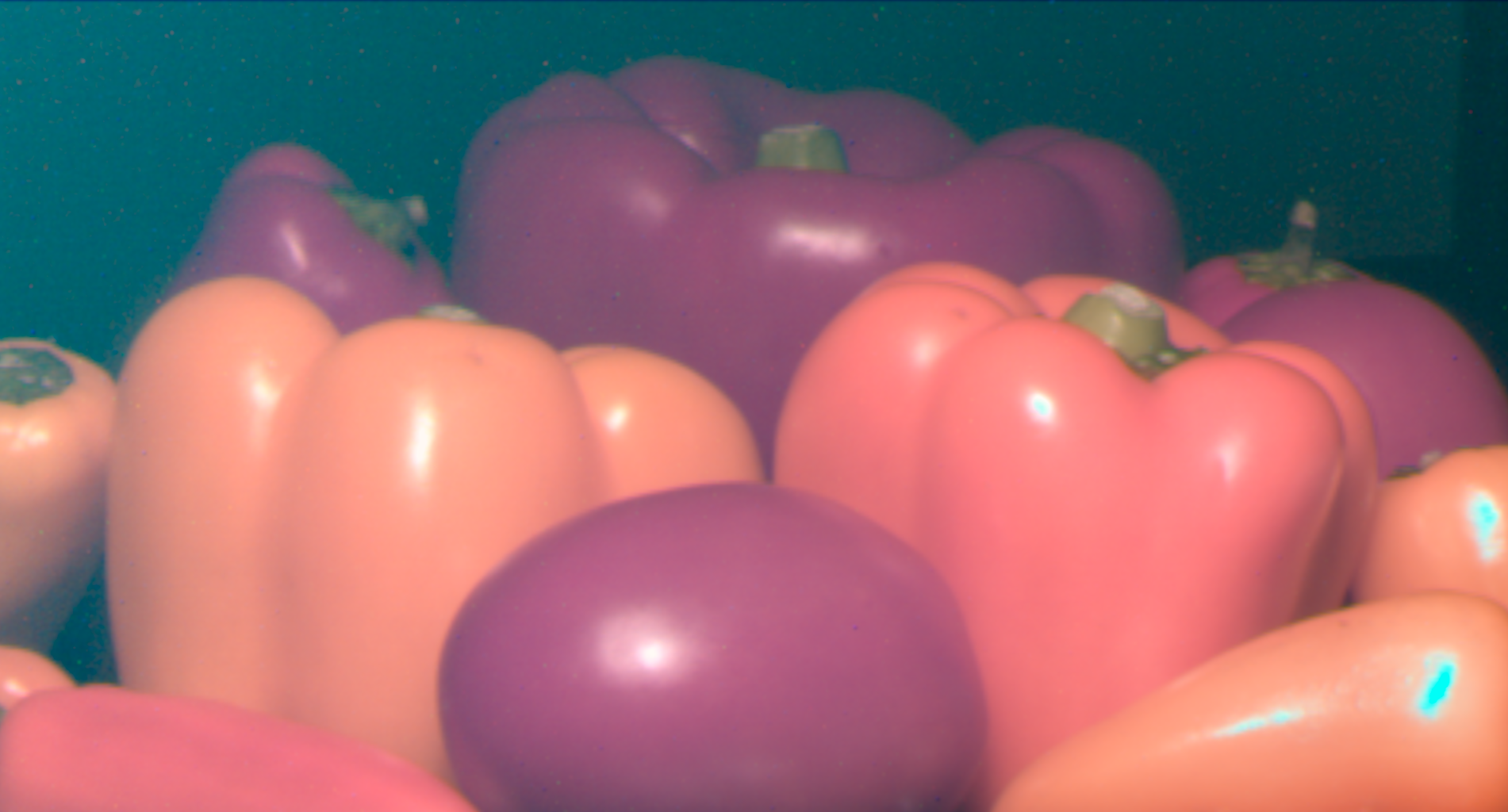}}\
  \subfloat[Our Network]{\includegraphics[width=0.49\columnwidth]{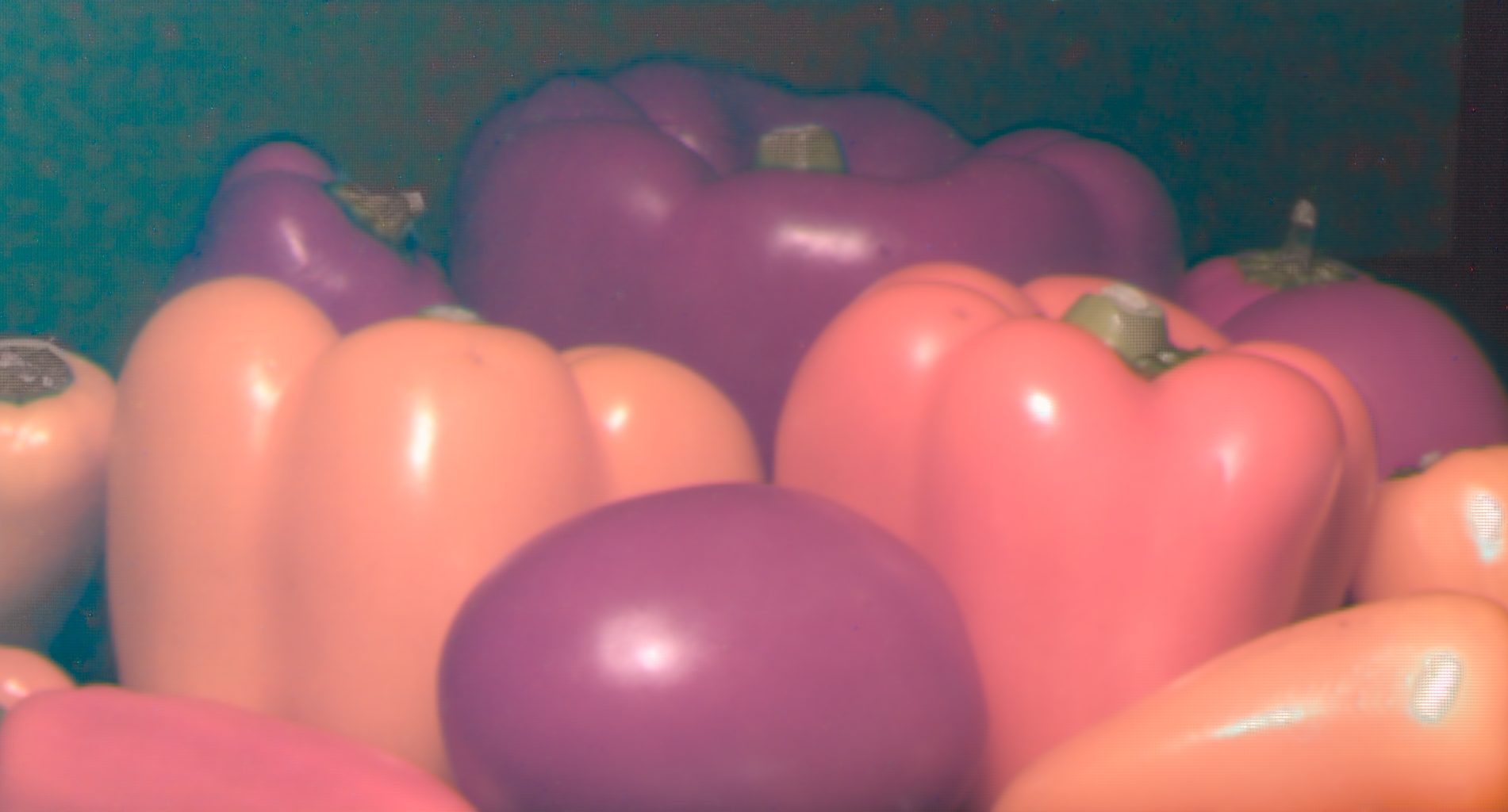}}
  \caption{RGB representation of one captured HSI scene as (a) high resolution ground truth image
  and (b) demosaiced images using our network.}
	\label{fig:resVeg2_entire}
\end{figure}
In the following, we visually analyze of differences between individual result images. This qualitative analysis includes a spectral signature analysis, analyzing the spectral behavior of the predicted hyperspectral cube across all 16 wavelengths. Additionally, we provide a visual comparison for the single spectral bands as well as calculated RGB image of the predicted hyperspectral cube. Fig.~\ref{fig:resVeg2_entire} shows full-resolution RGB images (
ground truth image and demosaiced result with our approach) of one example image in our captured dataset (dimension of $1024\times 2048$ pixels). The RGB images have been calculated using the CIE color matching functions \cite{stockman2019cone} with standard illuminant D65. Note, that the calculated RGB images show missing red components, since the snapshot camera is only sensitive up to $638$ nm and thus information is missing in the red channel of the RGB images.

\begin{figure}[tb]
  \centering
  \subfloat[Original Input]{\includegraphics[width=0.248\columnwidth]{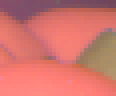}}\
  \subfloat[Ground Truth]{\includegraphics[width=0.25\columnwidth]{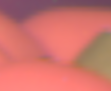}}\
  \subfloat[Our Network]{\includegraphics[width=0.25\columnwidth]{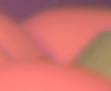}}
  \\
  \subfloat[DeepCND \cite{habtegebrial2019deep}]{\includegraphics[width=0.244\columnwidth]{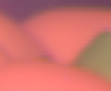}}\
  \subfloat[DCCNN \cite{dijkstra2019hyperspectral}]{\includegraphics[width=0.244\columnwidth]{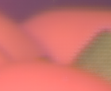}}\
  \subfloat[Lin. Interpolation]{\includegraphics[width=0.244\columnwidth]{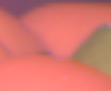}}\
  \subfloat[Cubic Interpol.]{\includegraphics[width=0.244\columnwidth]{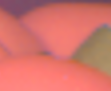}}
  \caption{RGB representation of a crop of Fig.~\ref{fig:resVeg2_entire} (center of the right image half). Our result is most similar to the ground truth, while traditional interpolation shows color artifacts at the edges.}
	\label{fig:resVeg2_zoom}
\end{figure}
Fig.~\ref{fig:resVeg2_zoom} shows enlarged cutouts of the scene depicted in Fig.~\ref{fig:resVeg2_entire} for the ground truth images as well as the input image and demosaiced versions of our network as well as linear and bilinear interpolation and the reference demosaicing networks. 
Compared to the original input image, our demosaiced result shows more structure while maintaining edges and color. In the reference networks the mosaic pattern is visible over the entire demosaiced image, c.f.~Fig.~\ref{fig:resVeg2_zoom}(d) and (e). This effect is drastically reduced by our approach, where this artifact only slightly appears especially around specular reflections as well as some homogeneous but noisy regions, which are not perfectly in focus, c.f. Fig.~\ref{fig:resVeg2_zoom}(c). Classical linear and cubic interpolation increase noise present in the input images, while single-pixel information is lost resulting in not correctly reconstructed edges. For the interpolation-based results, color artifacts appear around the edges (Fig.~\ref{fig:resVeg2_zoom}(f) and (g)) due to wrong spectral demosaicing. On the other hand, all neural network approaches learned to denoise the image as well as to correct the crosstalk.

\begin{figure}[t]
  \centering
  \subfloat[Homogeneous Area]{\includegraphics[width=0.45\columnwidth]{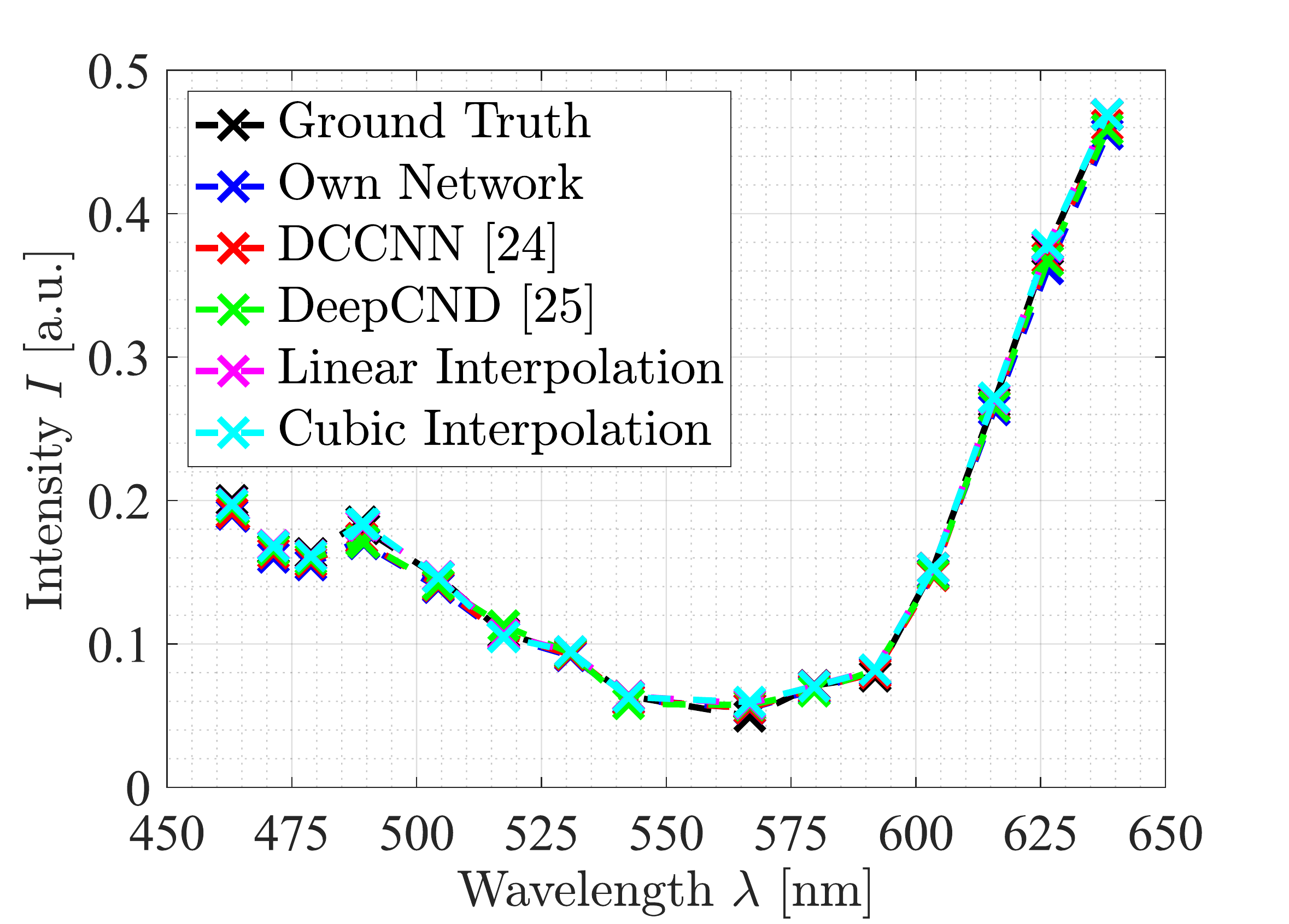}}
  \subfloat[Edge]{\includegraphics[width=0.45\columnwidth]{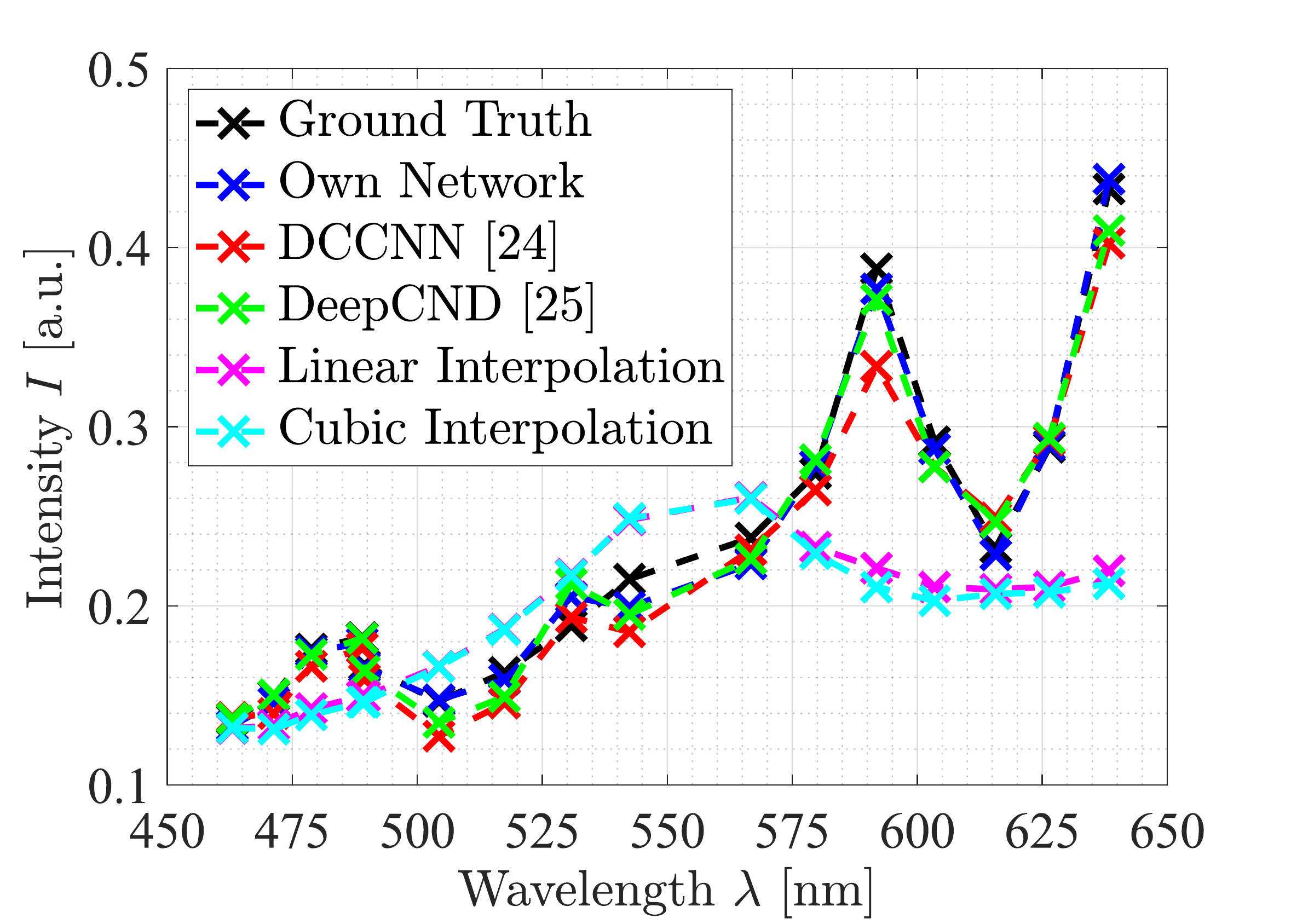}}
  \caption{Spectral plots of (a) a homogeneous area of the centered backmost red pepper under the total reflection and (b) an edge of the green and yellow parts visible in Fig.~\ref{fig:resVeg2_zoom}.}
	\label{plot:resVeg2_spectral}
\end{figure}

Fig.~\ref{plot:resVeg2_spectral} plots the spectral signatures of two example regions in the image for all compared demosaicing approaches. For homogeneous regions, the spectral signatures of all analyzed methods are close to that of the ground truth image region, c.f.~Fig~\ref{plot:resVeg2_spectral}(a). For edges, the spectral signature produced by our network almost completely follows that of the ground truth, followed closely by DeepCND \cite{habtegebrial2019deep} and then DCCNN \cite{dijkstra2019hyperspectral}, while the classical interpolation methods are not able to reconstruct similar characteristics of the spectral curves, resulting in wrong color appearance in the RGB images.

\subsection{Intraoperative Image Demosaicing}
\begin{figure}[ht]
  \centering
  \includegraphics[width=0.7\columnwidth]{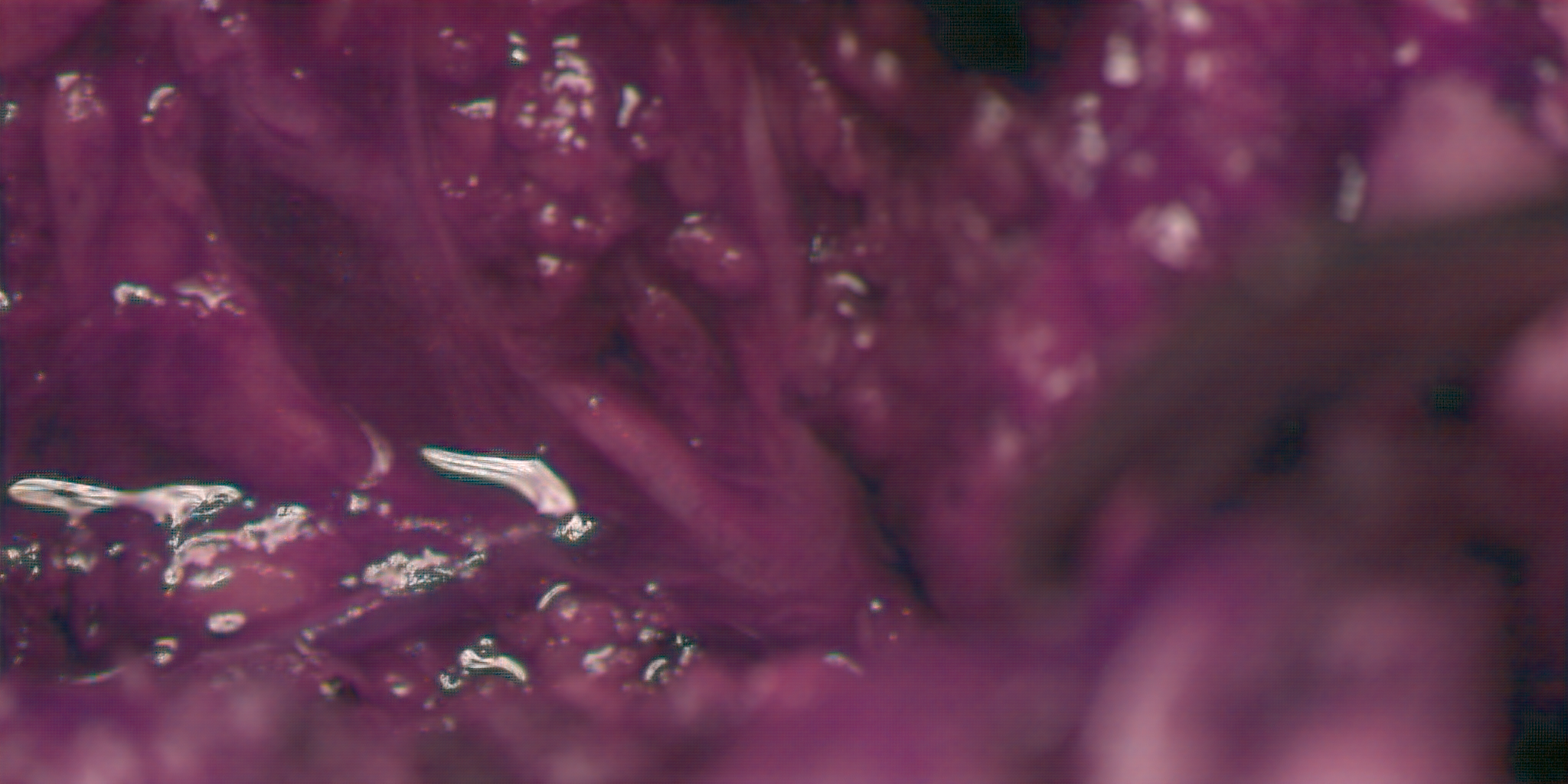}
  \caption{RGB representation of a captured surgical scene using the $4\times 4$ snapshot camera and demosaiced with our network.}
	\label{fig:resMed_entire}
\end{figure}

\begin{figure}[t]
  \centering
  \subfloat[Original Input]{\includegraphics[width=0.3\columnwidth]{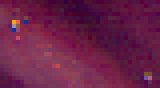}}\
  \subfloat[Our Network]{\includegraphics[width=0.3\columnwidth]{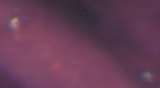}}\
  \subfloat[DeepCND \cite{habtegebrial2019deep}]{\includegraphics[width=0.3\columnwidth]{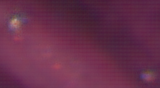}}\\
  \subfloat[DCCNN \cite{dijkstra2019hyperspectral}]{\includegraphics[width=0.3\columnwidth]{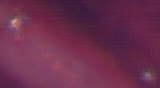}}\
  \subfloat[Linear Interpolation]{\includegraphics[width=0.3\columnwidth]{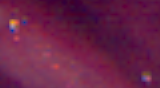}}\
  \subfloat[Cubic Interpolation]{\includegraphics[width=0.3\columnwidth]{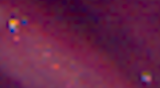}}
  \caption{RGB represenation of a crop of Fig.~\ref{fig:resMed_entire} (image center).}
	\label{fig:resMed_zoom}
\end{figure}

We also applied our approach to intraoperative image data, as HSI can be used in intraoperative settings in order to differentiate between different tissue types \cite{Wisotzky2020JMI} or to extract vital information \cite{WisotzkyComparision2021,Kossack_2022_CVPR}. Fig.~\ref{fig:resMed_entire} shows an example of a demosaiced image acquired intraoperatively. Our network is able to demosaic the original input and perform denoising as well as crosstalk correction at the same time, while the overall texture and color appearance is preserved. In the overall impression, all visual analyses from the test data are confirmed, c.f.~Fig.~\ref{fig:resMed_zoom}.

\section{Discussion and Conclusion}
In this work, we propose a neural network architecture for hyperspectral demosaicing for snapshot $4\times 4$ mosaic cameras. 
Additionally, we present a new unprecendened ground-truth dataset of real scenes for training, acquired with a sensor shifting unit and pixel reordering strategy. This dataset is combined with synthetic data generated from the CAVE  \cite{yasuma2010generalized} dataset. The performance of the network is evaluated using SSIM and PSNR scores and compared to traditional interpolation methods as well as to two demosaicing networks \cite{dijkstra2019hyperspectral,habtegebrial2019deep}. The results show that our proposed network with parallel feature extraction outperforms the reference networks with nearly $2\%$ and $0.4\%$ increase in SSIM score and $1.05$ dB and $0.6$ dB increase in PSNR. The scores of the classical interpolation methods are lowest as no spectral-spatial correlations are taken into consideration during demosaicing, through which they fail to preserve the spectral signature. This behavior visually emerges at edges. 
All network-based approaches are able to preserve the spectral signatures.

In detail, the results show that increasing the number of filters allows the network to learn more features (e.g., spatial-spectral correlations and edges) resulting in improved demosaicing. The improved performance for our network can be attributed to the feature addition layer, which combines the features from the M2C layer and the deep features extracted through the four residual blocks from the hand-crafted M2C input. This indicates that the upsampling performed on extracted feature maps (from four residual blocks) might yield better spatial-spectral resolution in the full spectrum mosaic cube. Currently, our network is only designed for $4\times 4$ mosaic pattern but similar architectures are possible for $3\times 3$ or $5\times 5$ patterns through adaptions at the deconvolutional layers.

Visual inspection of the results show a slight presence of the $4\times 4$ mosaic pattern in the demosaiced images, due to the challenge of creating a real world dataset using the 1-pixel shifting of a captured 3D scene. This 1-pixel shifting is dependent on the distance between camera and object, while the 3D nature of the scene did not allow an exact 1-pixel shift for the entire scene leaving a shadowing mosaic pattern as artifact. One approach to decrease or overcome this problem would be to use synthetic data, which would be completely aligned to (specific) mosaic snapshot cameras. We assume that this will further improve the results, since demosaicing results are heavily dependent on the quality of the full spectrum hyperspectral data available for training.

\section*{Acknowledgment}
This work was funded by the German Federal Ministry of Education and Research (BMBF) under Grant No.~16SV8061 (MultiARC) and the German Federal Ministry for Economic Affairs and Climate Action (BMWi) under Grant No.~01MK21003 (NaLamKI). Only tissue that has been exposed during normal surgical treatment has been scanned additionally with our described camera. This procedure has been approved by Charit\'e--Universit\"atsmedizin Berlin, Germany.

%
%
%
%
\bibliographystyle{splncs04}
\bibliography{literature}

\end{document}